\documentclass[letterpaper]{article} % DO NOT CHANGE THIS
\usepackage{aaai25}  % DO NOT CHANGE THIS
\usepackage{times}  % DO NOT CHANGE THIS
\usepackage{helvet}  % DO NOT CHANGE THIS
\usepackage{courier}  % DO NOT CHANGE THIS
\usepackage[hyphens]{url}  % DO NOT CHANGE THIS
\usepackage{graphicx} % DO NOT CHANGE THIS
\usepackage{natbib}  % DO NOT CHANGE THIS AND DO NOT ADD ANY OPTIONS TO IT
\usepackage{caption} % DO NOT CHANGE THIS AND DO NOT ADD ANY OPTIONS TO IT
\usepackage{algorithm}
\usepackage{algorithmic}

\usepackage{multirow}
\usepackage{color}
\usepackage{amsmath}
\usepackage{xcolor}
\usepackage{listings, xcolor}
\usepackage{colortbl}
\usepackage{xspace}
\usepackage{enumitem}
\usepackage{booktabs}
\usepackage{bbding}
\usepackage{pifont}
\usepackage{newfloat}
\usepackage{listings}
\usepackage{amsfonts}
\newcommand{\model}[0]{{\textsc{Amar }\xspace}} % Amar means love in Spanish
\usepackage{newfloat}
\usepackage{listings}
\DeclareCaptionStyle{ruled}{labelfont=normalfont,labelsep=colon,strut=off} % DO NOT CHANGE THIS
\floatstyle{ruled}
\newfloat{listing}{tb}{lst}{}
\floatname{listing}{Listing}
\title{Harnessing Large Language Models for Knowledge Graph Question Answering
via Adaptive Multi-Aspect Retrieval-Augmentation}
\author{
    Derong Xu\textsuperscript{\rm 12}, Xinhang Li\textsuperscript{\rm 1}, Ziheng Zhang\textsuperscript{\rm 3}, Zhenxi Lin\textsuperscript{\rm 3}, Zhihong Zhu\textsuperscript{\rm 4}, Zhi Zheng\textsuperscript{\rm 1}, Xian Wu\textsuperscript{\rm 3}\thanks{Corresponding authors.}, Xiangyu Zhao\textsuperscript{\rm 2}\footnotemark[1], Tong Xu\textsuperscript{\rm 1}\footnotemark[1], Enhong Chen\textsuperscript{\rm 1}
}
\affiliations{
\textsuperscript{\rm 1}University of Science and Technology of China \& State Key Laboratory of Cognitive Intelligence,\\ \textsuperscript{2}City University of Hong Kong, \textsuperscript{3}Jarvis Research Center, Tencent YouTu Lab,
\textsuperscript{4}Peking University\\
\{derongxu,xinhangli,zhengzhi97\}@mail.ustc.edu.cn, \{zihengzhang, chalerislin, kevinxwu\}@tencent.com, xianzhao@cityu.edu.hk, \{tongxu, cheneh\}@ustc.edu.cn, zhihongzhu@stu.pku.edu.cn

}

\begin{document}

\maketitle

\begin{abstract}
Large Language Models (LLMs) demonstrate remarkable capabilities, yet struggle with hallucination and outdated knowledge when tasked with complex knowledge reasoning, resulting in factually incorrect outputs. Previous studies have attempted to mitigate it by retrieving factual knowledge from large-scale knowledge graphs (KGs) to assist LLMs in logical reasoning and prediction of answers. However, this kind of approach often introduces noise and irrelevant data, especially in situations with extensive context from multiple knowledge aspects. In this way, LLM attention can be potentially mislead from question and relevant information. In our study, we introduce an \underline{A}daptive \underline{M}ulti-\underline{A}spect \underline{R}etrieval-augmented over KGs (\textsc{Amar}) framework. This method retrieves knowledge including entities, relations, and subgraphs, and converts each piece of retrieved text into prompt embeddings. The \textsc{Amar} framework comprises two key sub-components: 1) a self-alignment module that aligns commonalities among entities, relations, and subgraphs to enhance retrieved text, thereby reducing noise interference; 2) a relevance gating module that employs a soft gate to learn the relevance score between question and multi-aspect retrieved data, to determine which information should be used to enhance LLMs' output, or even filtered altogether. Our method has achieved state-of-the-art performance on two common datasets, WebQSP and CWQ, showing a 1.9\% improvement in accuracy over its best competitor and a 6.6\% improvement in logical form generation over a method that directly uses retrieved text as context prompts. These results demonstrate the effectiveness of \textsc{Amar} in improving the reasoning of LLMs.
\begin{links}
\link{Code}{https://github.com/Applied-Machine-Learning-Lab/AMAR}
\end{links}

\end{abstract}
\section{Introduction} \label{sec:intro}

Recently, large language models (LLMs) like GPT-4~\cite{openai2024gpt4} and Llama~\cite{touvron2023llama} have shown impressive performance improvements across a variety of natural language processing (NLP) tasks \cite{LeiWANG186345,MuningWEN176349,DerongXU186357,YufeiZENG176340,wu2024survey,liu2024unimel}. However, when dealing with specialized knowledge not in training corpus and complex knowledge reasoning, LLMs still struggle with outdated knowledge and hallucination problems \cite{ji2023survey}. They thus may produce factually incorrect outputs, limiting their usefulness in the areas requiring high reliability, such as healthcare \cite{he2023survey,liu2024moe} and safety \cite{dong2024attacks}. 
 % which have garnered considerable attention from researchers to explore their capabilities for complex knowledge reasoning \cite{TOG}
To solve complex reasoning with specialized knowledge, a line of research \cite{chatkbqa,TOG} have explored Knowledge Graph Question Answering (KGQA), a task that improves logical reasoning and prediction of answers by retrieving reliable information from large-scale knowledge graphs (KGs) like Freebase~\cite{bollacker2008freebase} and Wikidata~\cite{wikidata}. KGs store extensive factual knowledge in a structured format known as triplets \cite{xu2024multi,xu2022relation,wang2023federated}, consisting of \textit{(head entity, relation, tail entity)}, which is seen as a potential solution for enhancing the interpretability of LLMs reasoning \cite{TOG}.
\begin{figure}[t]
\centering
\includegraphics[width=0.9\linewidth]{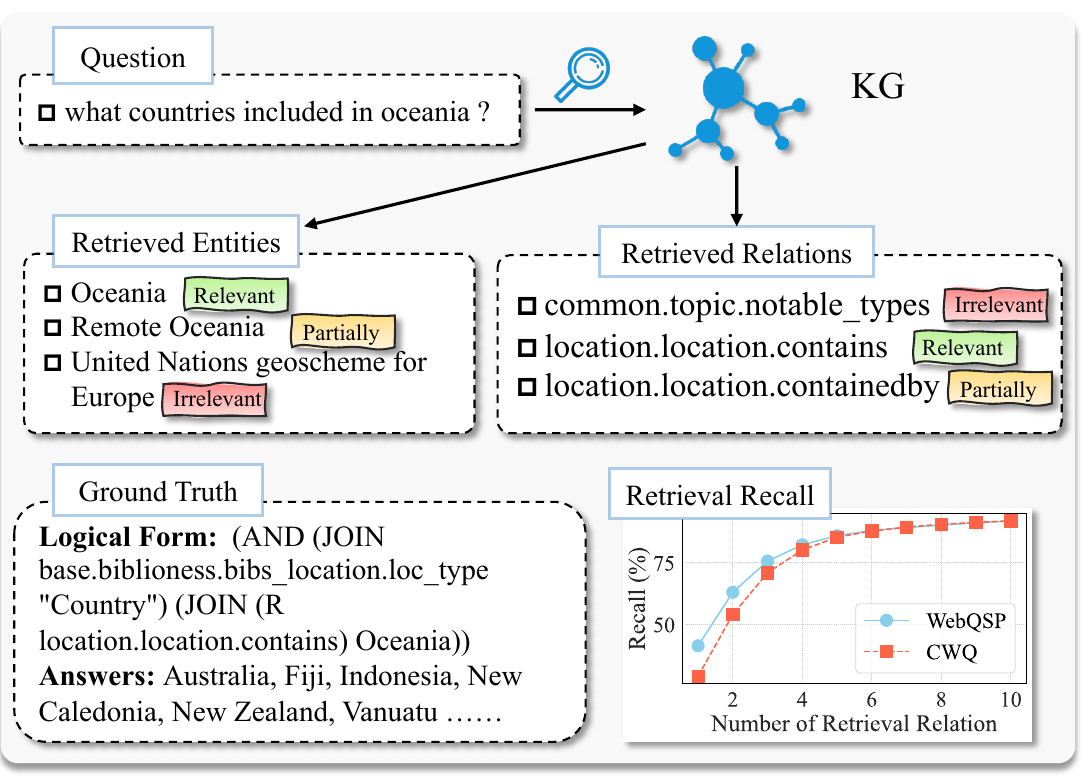}
\caption{ An example illustrates the retrieval results categorized as \textit{relevant}, \textit{irrelevant}, and \textit{partially relevant}.
}
\label{fig:intro}
\end{figure}

The recent advancements in KGQA can be broadly classified into two main categories.
The first category involves using LLMs to transform input questions into structured logical forms~\cite{chatkbqa}, such as S-expressions, which can then be queried on KG using a graph database query language like SPARQL to obtain the final answers. These approaches leverage the structured nature of KG to extract precise information in response to complex queries.
The second category of KGQA involves directly predicting answers using LLMs without extra logical forms. Some works in this category perform multi-hop reasoning on KG in a step-by-step manner by querying LLMs \cite{TOG}. Other works retrieve additional contextual information related to the question and use it as contextual input to enhance LLMs~\cite{G-retriever,RoG}. These approaches aim to improve the accuracy of answers by incorporating more context into the reasoning process.

However, the retrieval process often yields a large amount of information, as shown in Figure~\ref{fig:intro}, which may be irrelevant or only partially relevant to the question and ground truth. The retrieval recall curve demonstrates a positive correlation between the number of retrieved data and the recall score, indicating that the retrieved data contain valuable information.
Therefore, it raises a question: \textit{How to identify which retrieved knowledge is valuable and which is not?} This question becomes particularly crucial when a significant portion of retrieved knowledge does not provide substantial assistance. In terms of this, we observed that previous studies \cite{decaf,GMT-KBQA} failed to considering commonalities among different aspects of retrieved information, which can be beneficial in identifying crucial knowledge. And they also neglected adaptive learning to establish relevance between question and retrieved text.
 
To address these overlooked challenges, we propose an \underline{a}daptive \underline{m}ulti-\underline{a}spect \underline{r}etrieval-augmented over KG (\textsc{Amar}) framework. Instead of directly appending the retrieved information as context input, \model utilizes the retrieved information more flexibly, therefore enhancing LLMs reasoning.
\model primarily includes two modules:
 1) Self-alignment module, where multi-aspect retrieval data (including entities, relations, and subgraphs, all of which are linearized as text) is separately mapped into prompt embeddings from text, facilitating fine-grained tuning. Next, cross-attention and self-attention are applied to the multi-aspect embeddings to obtain consistency tokens. The aim is to align the commonalities among different pieces of information. For instance, if an entity and a subgraph both mention ``\textit{Oceania}'', this implies that they are probably consistent in conveying the same piece of information. In this way, we can enhance crucial knowledge, thereby reducing noise interference.
 2) Relevance gating module, which measures the relevance between the embedding of question and consistency tokens. This module introduces siamese networks with shared weights to learn relevance score, which serves as a soft gate to adaptively decide which retrieved information is more useful for LLMs reasoning.
Through the self-alignment and relevance gating modules, \model adaptively filters and selects multi-aspect retrieval knowledge, enabling a more rational utilization of context and avoiding interference from noise. We then fine-tune LLM in a parameter-efficient manner to leverage the retrieved knowledge, assisting in generating reasonable logical forms. Finally, we refine the logical forms using the similarity of entities and relations and query KG to obtain final answers.
Overall, this work makes three key contributions:
\begin{itemize}%[topsep=0pt, partopsep=0pt, leftmargin=13pt, parsep=0pt, itemsep=3pt]
    \item To the best of our knowledge, this is the first study to enhance LLMs reasoning for KGQA tasks by utilizing multi-aspect KG information as prompt embeddings. 
    \item  We propose novel self-alignment and relevance gating modules, which enable LLMs to adaptively filter and select multi-aspect retrieval knowledge, allowing for more rational utilization of context while effectively avoiding interference from noise.
    \item The effectiveness of \model was extensively validated on two datasets across five metrics. \model demonstrated superior performance, achieving state-of-the-art (SOTA) results compared to 22 strong baselines.
\end{itemize}

\section{Preliminaries} \label{sec:preliminaries}
We first introduce key concepts and notations for our task.

\textbf{Knowledge Graph (KG)}. A KG is a collection of factual knowledge organized in form of triples: $\mathcal{G} =  \{(h,r,t)\} \subseteq \mathcal{E} \times \mathcal{R} \times  \mathcal{E} $, where $\mathcal{E}$ represents the entity set and $\mathcal{R}$ represents relation set. Each triple consists of three elements: a head entity $h$, a relation $r$, and a tail entity or a literal $t$.

\textbf{Logical Form}. The logical form is a structured language representation of a question. In this work, we adopt S-expression $\mathcal{F}$ as our chosen logical expression, following \cite{chatkbqa,decaf}. As shown in the examples provided in Figure \ref{fig:intro}, the S-expression utilizes functions (such as \textit{JOIN}, \textit{AND}) that operate on set-based semantics, which keep a balance between readability and compactness and thus is well-suited for KGQA \cite{gu2021beyond}.

\textbf{KGQA with LLM}. KGQA is a classical NLP task that has been further enhanced through the utilization of LLMs.
Given a question $q$, this task aims to retrieve knowledge related to $q$ from a KG $\mathcal{G}$ and generate an S-expression $\mathcal{F}$. Since many KG storage engines support SPARQL, the generated S-expression $\mathcal{F}$ is converted into a SPARQL query, which is further executed against $\mathcal{G}$ to obtain the final answers. In our work, We retrieve multi-aspect knowledge (including entities $k_e$, relations $k_r$, and subgraphs $k_s$) and design a model $f$ that utilizes LLMs to generate $\mathcal{F}$ based on question and retrieval knowledge as input, i.e., $\mathcal{F} = f(q,k_e,k_r,k_s)$. By converting and executing the SPARQL query on the KG, we can obtain final answers denoted as $a = query(convert(\mathcal{F})) \in \mathcal{A}_q$, where $\mathcal{A}_q \subseteq \mathcal{E}$.

\section{Methodology} \label{sec:method}
\begin{figure*}[t]
\centering
\includegraphics[width=0.9\linewidth]{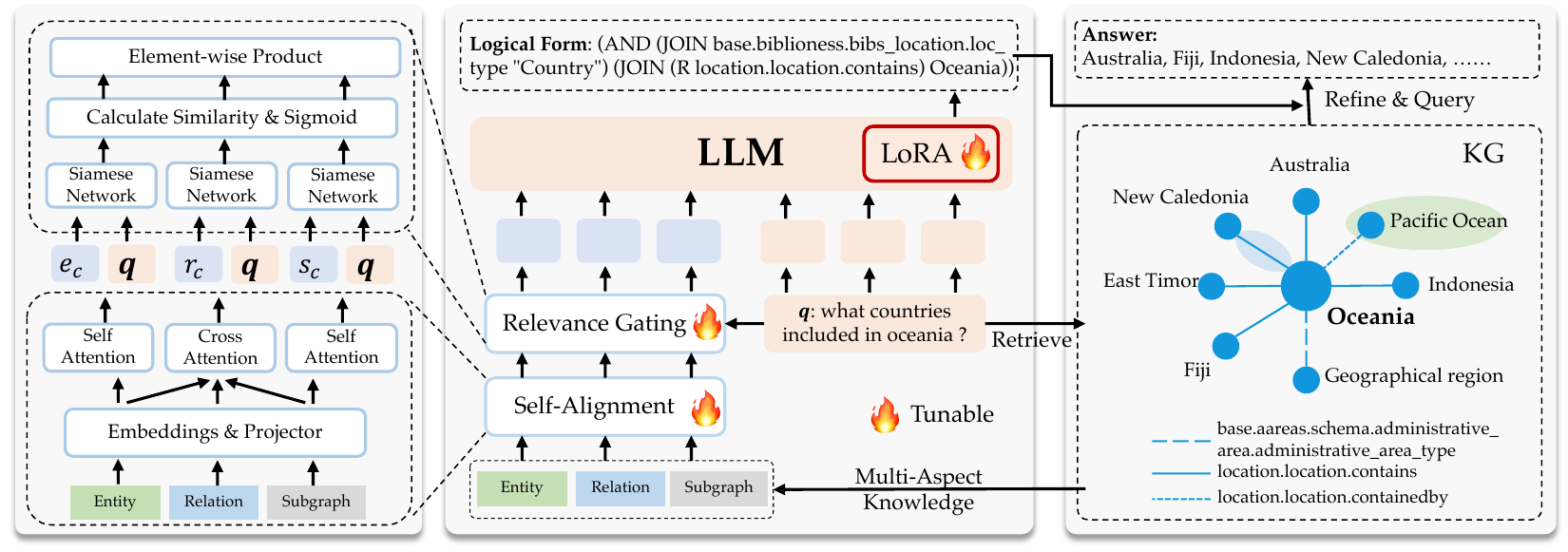}
\caption{Overall framework of \model. We retrieve multi-aspect knowledge from KG and obtain weighted consistency prompt embeddings using Self-Alignment and Relevance Gating modules. These embeddings are combined with the question and fed into LLMs. The generated logical expression is refined and used to query KG, ultimately getting final answers. The unlabeled color blocks in the middle represent the tokens input to the LLM, with pink blocks denoting question tokens and blue blocks denoting retrieved knowledge tokens.
}
\label{fig:method}
\end{figure*}

In this section, as shown in Figure \ref{fig:method}, we begin with multi-aspect knowledge retrieval. We then delve into two specific modules, providing detailed explanations of their functionalities, and demonstrate the process of querying KG using logical form.

\subsection{Retrieval of Multi-Aspect Knowledge} \label{sec:Retrieve Multi-level Knowledge}
Previous KGQA methods either focus on retrieving entities and relations \cite{GMT-KBQA} or retrieving multiple triplets to form subgraphs \cite{decaf,G-retriever}. 
In our research, we propose to leverage three types of knowledge simultaneously. Each possesses varying aspects of information, which complement each other and also share commonalities. This enables us to effectively align crucial knowledge across three types.
Due to the differences in their representations (e.g., length and format), we employ different retrieval methods for each type.

\textbf{Entity Retrieval.} 
One effective approach for retrieving candidate entities $k_e$ is to conduct entity linking with question $q$. Following \citet{GMT-KBQA}, we employ the ELQ \cite{ELQ} for question entity linking, which utilizes a bi-encoder to simultaneously detect mentions and link them to entities end-to-end. Then FACC1 \cite{facc1} (a comprehensive Freebase annotation of corpora) is employed to identify entities that were not linked by ELQ, to enhance the range of candidate entities.

\textbf{Relation Retrieval.}
In large-scale KG (e.g. Freebase), relations are typically organized hierarchically, such as the example \textit{base.biblioness.bibs\_location.loc\_type}. Therefore, directly using question-based dense retrieval for similarity may not be effective. To address this, following \citet{GMT-KBQA}, we propose masking entity mentions detected during the candidate entity retrieval stage with a [BLANK] token for each question $q$. Building on the work of \citet{GMT-KBQA,CBR-KBQA}, we train two separate BERT models that encode questions and relations into a shared dense space. The objective of optimization is to maximize the score of the relevant relation compared to randomly sampled relations. To retrieve the nearest relations, we employ FAISS \cite{douze2024faiss}, a highly efficient vector database, which allows us to speed up the search process and obtain the most relevant results. 

\textbf{Subgraph Retrieval.}
One crucial consideration is the wealth of structural and semantic information contained within KG. 
Since KG data is typically stored as triplets, we linearize triplets by combining head entity, relation, and tail entity for retrieval. Following \cite{decaf}, we propose grouping linearized sentences with the same head entity into a document. To save computing resources, we only focus on 1-hop subgraphs to capture structural information.
Furthermore, concerning the potential information loss when converting long documents into vectors, we employ sparse retrieval approaches that rely on keyword dependencies. Specifically, we employ techniques like BM25, which calculates TF-IDF scores based on sparse word matches between input questions and KB-linearized passages. For more information, refer to the appendix.

\subsection{Self-Alignment} \label{sec:Self-Alignment}
Although multi-aspect retrieval offers comprehensive auxiliary information, it can also introduce irrelevant knowledge and noise, resulting in negative impacts. To address this concern, we propose to align the commonalities among multi-aspect information to improve informativeness.

We first utilize LLM's embedding layer to convert the multi-aspect retrieval knowledge into text embeddings $\boldsymbol{X}_{e} \in \mathbb{R}^{t_k \times l \times e}$, where the $t_k$ represent top $k$ retrieval texts, $l$ denotes the maximum length of text, and $e$ indicate the embedding dim of the token. We then apply an averaging operation, followed by a projector network \(\mathcal{M}\), which transforms these embeddings into prompt embeddings $\boldsymbol{X}_{t} \in \mathbb{R}^{t_k \times e}$ (\textit{i.e.}, one piece of retrieval text is projected to one token embedding). For the sake of efficiency, $\mathcal{M}$ is designed to consist of down-projection and up-projection layers, with a nonlinear layer situated between them, as follows:

\begin{equation}
\begin{aligned}
    &\boldsymbol{X}_{e} = Embeddings({T}), \\
    &\boldsymbol{X}_{t} = \mathcal{M}(\hat{\boldsymbol{X}}_{e}), \quad \hat{\boldsymbol{X}}_{e} =  \frac{1}{l}\sum\nolimits_{i=1}^{l} \boldsymbol{X}_e[:,i,:],\\
\end{aligned}
\end{equation}
where $T$ represents the retrieval text of entities, relations, or subgraphs.
Next, we apply self-attention separately to the prompt embeddings of entities and relations to obtain entity-consistency tokens $\boldsymbol{e}_{c} $ and relation-consistency tokens $\boldsymbol{r}_{c} $:
\begin{equation}
\begin{aligned}
    &\boldsymbol{e}_{c} = Self\text{-}Attn(\boldsymbol{E}_{t}) \in \mathbb{R}^{t_k \times e}, \\
   &\boldsymbol{r}_{c} = Self\text{-}Attn(\boldsymbol{R}_{t}) \in \mathbb{R}^{t_k \times e}, \\
\end{aligned}
\end{equation}

Here, we use $\boldsymbol{S}_{t}$, $\boldsymbol{E}_{t}$, and $\boldsymbol{R}_{t}$ represent prompt embeddings $\boldsymbol{X}_{t}$ of subgraphs, entities, and relations, respectively.
This attention is applied to individual retrieval text (e.g., one sentence) rather than individual tokens to learn the correlation and consistency between different retrieval information and their importance within the entire top $k$ retrieval data.

In addition, we further focus on leveraging subgraphs information. To determine the significance of retrieval text, we employ entities and relations as alignment factors. As shown in Figure \ref{fig:method}, by aligning triples in a subgraph that is highly consistent to relation  `\textit{location.location.contains}', we can weight important data through commonalities of multi-aspect knowledge. To obtain subgraph-consistency tokens $s_{c}$, we perform cross-attention to entity-subgraph and relation-subgraph pairs, respectively:
\begin{equation}
\begin{aligned}
    &\boldsymbol{s}_{c}^{e} = Cross\text{-}Attn(\boldsymbol{S}_{t}, \boldsymbol{E}_{t} ,\boldsymbol{E}_{t}) \in \mathbb{R}^{t_k \times e}, \\
   &\boldsymbol{s}_{c}^{r} = Cross\text{-}Attn(\boldsymbol{S}_{t}, \boldsymbol{R}_{t} ,\boldsymbol{R}_{t}) \in \mathbb{R}^{t_k \times e}, \\
\end{aligned}
\end{equation}
and sum the results to get $\boldsymbol{s}_{c}= \boldsymbol{s}_{c}^{e} + \boldsymbol{s}_{c}^{r} \in \mathbb{R}^{t_k \times e}$. The consistency tokens of entity $\boldsymbol{e}_{c}$, relation $\boldsymbol{r}_{c}$, and subgraph $\boldsymbol{s}_{c}$ contain refined knowledge aligned between multi-aspect information, enhancing the utilization of the retrieved data.

\subsection{Relevance Gating} \label{sec:Relevance Gating}
After obtaining the consistency tokens, we expect the model to learn the relevance between retrieval data and the question.  The relevance is used to construct a soft gating mechanism, which will adaptively select the relevant retrieval information to be utilized.
To achieve this, we design a siamese network for each type of consistency token to measure its relevance to the question embedding $\boldsymbol{Q}_{e} \in \mathbb{R}^{l \times e}$. Each siamese network consists of a shared MLP $\mathcal{M}_{share}$ networks that processes both question embedding and consistency tokens, the generated $\boldsymbol{q}_{m}\in \mathbb{R}^{l \times e}$ and $\boldsymbol{x}_{c}\in \mathbb{R}^{t_k \times e}$ are then used to calculate similarity score $\boldsymbol{G}_{sim} \in \mathbb{R}^{t_k \times l}$ through a batch matrix multiplication. This score is subsequently averaged, and a sigmoid activation function is applied to produce the final relevance score $\boldsymbol{g}$, as described below:
\begin{equation}
\begin{aligned}
    &\boldsymbol{q}_{m} = \mathcal{M}_{share}(\boldsymbol{Q}_{e}), \quad \boldsymbol{x}_{c} = \mathcal{M}_{share}(\boldsymbol{X}_{c}), \\
    & \boldsymbol{g} = Sigmoid(\frac{1}{l}\sum\nolimits_{i=1}^{l} \boldsymbol{G}_{sim}[:,i]),\quad \boldsymbol{G}_{sim} = \boldsymbol{x}_{c} \cdot  \boldsymbol{q}_{m}^T.\\
\end{aligned}
\end{equation}

Here $\boldsymbol{X}_{c}$ can be denoted as entities, relations, or subgraphs consistency tokens.
The relevance score of entity $\boldsymbol{g}_{e}$, relations $\boldsymbol{g}_{r}$ and subgraph $\boldsymbol{g}_{s}$ serve as soft gates, are used to model the influence of each consistency tokens by element-wise product, respectively:
\begin{equation}
\begin{aligned}
&\boldsymbol{e}_{c}^{w} = \boldsymbol{g}_{e}\circ \boldsymbol{e}_{c}, 
&\boldsymbol{r}_{c}^{w} = \boldsymbol{g}_{r}\circ \boldsymbol{r}_{c},  \quad
&\boldsymbol{s}_{c}^{w} = \boldsymbol{g}_{s}\circ \boldsymbol{s}_{c}.
\end{aligned}
\end{equation}

To enhance generalization ability, we introduce randomly initialized soft tokens $\boldsymbol{p} \in \mathbb{R}^{l \times e}$, which are concatenated with the weighted consistency tokens and the question embedding. These combined embeddings are then fed into LLMs to generate logical form $\mathcal{F}$ as follows:

\begin{equation}
 % \mathcal{F} = \prod \limits_{i=1}^r p_{\theta,\phi_1,\phi_2}(y_i|y_{<i},[{e}_{c}^{w};{r}_{c}^{w};{s}_{c}^{w};{Q}_{e}]),
 \mathcal{F} = f_{\theta,\phi_1,\phi_2}([\boldsymbol{p};\boldsymbol{e}_{c}^{w};\boldsymbol{r}_{c}^{w};\boldsymbol{s}_{c}^{w};\boldsymbol{Q}_{e}]),
\end{equation}
where [;] represents concatenation operation, the parameters $\theta$ of LLMs itself are frozen. The parameters that require back-propagation optimization include our model parameters $\phi_1$ and LoRA parameters $\phi_2$.

\subsection{Query Execution} \label{sec:Query Execution}
Due to the long-tail distribution of fine-tuned data and the lack of specific knowledge, LLM may not strictly adhere to the contextual information provided \cite{chatkbqa}. As a result, the generated logical forms may contain non-existent entities or relations. For instance, when asked `\textit{where was rihanna born and raised?}', an LLM might generate the logical form `\textit{(JOIN (R people.person.place\_of\_birth) Rihana)}' instead of the correct spelling `\textit{Rihanna}'. This discrepancy renders the logical form non-executable on KG, and the same issue can also arise with relations.

To further refine the quality of entity and relation, we employ a similarity-based approach with KG. Specifically, we utilize an unsupervised SimCSE model to measure the similarity between each entity in the generated logical form $\mathcal{F}$ and the labels of entities in the entity set $\mathcal{E}$. By setting a threshold, we retain the most relevant entities $\mathcal{E}_{sub}$. Additionally, we query the KG to identify relations $\mathcal{R}_{2-hop}$ that are within 2 hops of the obtained subset of entities $\mathcal{E}_{sub}$. Similarly, we calculate the similarity between all relations in $\mathcal{F}$ and the identified set of relations. This refinement process allows us to generate a new list of candidate logical forms $\mathcal{F}_{new}$ that better align with the KG. After converting to SPARQL language, it can be used to query answers from KG: $a = query(convert(\mathcal{F}_{new}))$.

\section{Experiments} \label{sec:experiments}

\begin{table*}[t]
% \setlength\tabcolsep{3pt}  %可以控制列间距
% \renewcommand{\arraystretch}{1.1} %可以控制行间距
% \footnotesize
\centering
\fontsize{9}{11}\selectfont
\setlength{\tabcolsep}{2.6mm}{
\begin{tabular}{c|l|cccccc}
\toprule
\multicolumn{1}{c|}{\multirow{2}{*} {Type}} & \multicolumn{1}{l|}{\multirow{2}{*} {Model}}   & \multicolumn{3}{c}{WebQSP}   & \multicolumn{3}{c}{CWQ} \\
\cline{3-8}
 \multicolumn{1}{c|}{} & \multicolumn{1}{c|}{}   &F1&Hits@1& Acc &F1&Hits@1& Acc\\ 
\hline
\hline
\multicolumn{1}{c|}{\multirow{4}{*} {EM-based}} & KV-Mem {\scriptsize \cite{miller2016key}} & 34.5 & 46.7 & - & 15.7 & 18.4 & - \\
\multicolumn{1}{c|}{}& NSM$_{+h}$ {\scriptsize \cite{NSM}} &67.4 & 74.3& - & 44.0& 48.8&-   \\
\multicolumn{1}{c|}{}& TransferNet {\scriptsize \cite{shi2021transfernet}} &-  & 71.4 &- &- &48.6 &  - \\
\multicolumn{1}{c|}{}& KGT5 {\scriptsize \cite{KGT5}} &-  &56.1 &- &- & 36.5&  - \\
\hline
\multicolumn{1}{c|}{\multirow{5}{*} {IR-based}} &  GraftNet {\scriptsize \cite{sun-etal-2018-open}} & 60.4 & 66.4 & - & 32.7 & 36.8 & - \\
\multicolumn{1}{c|}{}& PullNet {\scriptsize \cite{sun-etal-2019-pullnet}} & - & 68.1 & - & - &45.9 & - \\
\multicolumn{1}{c|}{}& SR+NSM {\scriptsize \cite{zhang2022subgraph}} &  64.1 & 68.9 &- &47.1& 50.2&- \\
\multicolumn{1}{c|}{}& SR+NSM+E2E {\scriptsize \cite{zhang2022subgraph}} & 64.1  &69.5 &- &46.3&49.3 &-\\
\multicolumn{1}{c|}{}& UniKGQA {\scriptsize \cite{unikgqa}} & 71.0 &77.0& - &49.4& 50.9 &-\\
\hline
\multicolumn{1}{c|}{\multirow{6}{*} {SP-based}}& CBR-KBQA {\scriptsize \cite{CBR-KBQA}} & 72.8 &- &69.9 &70.0 &-& 67.1 \\
\multicolumn{1}{c|}{}& GMT-KBQA {\scriptsize \cite{GMT-KBQA}} & 76.6 &- &73.1 &77.0 &-& 72.2 \\
\multicolumn{1}{c|}{}& UnifiedSKG {\scriptsize \cite{xie-etal-2022-unifiedskg}} & 73.9 &-& -& 68.8& - &-\\
\multicolumn{1}{c|}{}& RnG-KBQA {\scriptsize \cite{rng-kbqa}} & 75.6&-&-&-&-&-\\
\multicolumn{1}{c|}{}& DecAF {\scriptsize \cite{decaf}} & 78.8&82.1&-&-&70.4&-\\
\multicolumn{1}{c|}{}& FC-KBQA {\scriptsize \cite{fc-kbqa}} & 76.9& - &- &56.4& -& -\\
\hline
\multicolumn{1}{c|}{\multirow{8}{*} {LLM-based}} & KD-CoT {\scriptsize \cite{wang2023knowledge}} & 52.5 & 68.6 & - &-  &55.7 & - \\
\multicolumn{1}{c|}{}& Pangu {\scriptsize \cite{pangu}} & 79.6 & - & - & - & - & - \\
\multicolumn{1}{c|}{}& StructGPT {\scriptsize \cite{structgpt}} & - & 72.6 & -  & -  & - & - \\
\multicolumn{1}{c|}{}& ChatKBQA {\scriptsize \cite{chatkbqa}} & 79.8 &83.2& 73.8 &77.8 &82.7 &73.3\\
\multicolumn{1}{c|}{}& ToG-R (GPT-4) {\scriptsize \cite{TOG}} & - & 82.6 & -  & -  & 69.5 & - \\
\multicolumn{1}{c|}{}& G-Retriever {\scriptsize \cite{G-retriever}} & -& 70.1& -& -& -& -\\
\multicolumn{1}{c|}{}& GNN-RAG {\scriptsize \cite{mavromatis2024gnn}} & 73.5 &82.8& -  &60.4& 62.8& - \\
\rowcolor{gray!10}  \multicolumn{1}{c|}{}& \model (Ours) & \textbf{81.2}\scriptsize{$\pm$0.15} & \textbf{84.3}\scriptsize{$\pm$0.16}  & \textbf{75.2}\scriptsize{$\pm$0.10} & \textbf{78.5}\scriptsize{$\pm$0.11} & \textbf{83.1}\scriptsize{$\pm$0.09} & \textbf{74.5}\scriptsize{$\pm$0.07} \\
\bottomrule
\end{tabular}}
\caption{Performance comparison of different types of KGQA methods on WebQSP and CWQ datasets. We present the Mean scores and standard deviations (mean ± std) of five experiments with different random seeds. The best result is highlighted in \textbf{bold}, and the baseline results are taken from corresponding papers.}
\label{tab:main result}
\end{table*}

\subsection{Experiment Settings}

\textbf{Datasets. }
Our experiments were conducted using two well-known datasets: WebQuestionsSP (WebQSP) \cite{webqsp} and ComplexWebQuestions (CWQ) \cite{CWQ}. The dataset statistics are presented in the Appendix. 
Both datasets contain SPARQL queries that correspond to the questions and can be executed on Freebase to obtain answers.

\noindent\textbf{Baselines.}
In this study, we evaluate performance with 22 baselines, which are categorized into four groups: embedding-based (EM-based), information retrieval-based (IR-based), semantic parsing-based (SP-based), and LLM-based methods. For more detailed descriptions of the baselines, please refer to the Appendix.
It is important to note that some methods, such as DecAF~\cite{decaf}, can be classified as multiple groups, specifically IR-based and SP-based. To ensure fairness, we do not include the results of using the oracle entity linking annotations setting, such as RoG \cite{RoG}. We put the performance comparison of the oracle setting in the Appendix.

\noindent\textbf{Evaluation Metrics.}
We use Hits@1, F1, and Acc as primary evaluation metrics following \cite{chatkbqa}. Hits@1 assesses the accuracy of top-1 predicted answer, F1 considers the coverage of all possible answers, and Acc measures the strict exact-match accuracy.
We further assess the quality of generated S-expressions by employing two metrics: the extract match ratio (EM) and the match after beam search ratio (BM) with ground-truth S-expressions, for analytical experiments.

\noindent\textbf{Implementation Details.}
% We employ LLaMA2-7B and LLaMA2-13B \cite{touvron2023llama} as LLM backbones, and then fine-tune LLMs using LoRA \cite{hu2022lora} on WebQSP and CWQ. 
Following ~\citet{chatkbqa}, we fine-tune LLaMA2-7B on WebQSP and LLaMA2-13B on CWQ using LoRA.
We evaluate the impact of backbones and fine-tuning methods in our subsequent experiments. During inference, we utilize beam search to generate multiple logical forms. We select the executable logical form with the highest score to obtain answers. All experiments were done on NVIDIA A6000 GPUs. We only searched the number of retrievals $k$ with values of \{4,8,16,32,64,100\}. 

\subsection{Main Results}
% In this section, we compare our method with all baselines.
As observed from Table \ref{tab:main result}, \model outperforms all baselines across all metrics on both datasets. Notably, on WebQSP dataset, accuracy has improved by 1.6\% compared to the second-best baseline, ChatKBQA, marking new state-of-the-art performance.
Specifically, \model surpasses subgraph retrieval techniques such as SR+NSM and DecAF with 26\% and 2.4\% F1 improvements on WebQSP, respectively, as well as entity and relation retrieval methods like GMT-KBQA with 5.3\% F1 improvement on WebQSP.
This can be attributed to our proposed self-alignment mechanism, which effectively aligns multi-aspect knowledge.
On the other hand, \model also outperforms other LLM-based approaches, such as G-Retriever and ChatKBQA, suggesting that our approach of learning prompt embeddings for retrieval can more flexibly leverage the capabilities of LLMs to utilize retrieval knowledge.

\begin{table}[t]
\setlength\tabcolsep{2pt}  %可以控制列间距
\footnotesize
\centering
\fontsize{9}{11}\selectfont
\setlength{\tabcolsep}{3mm}
\begin{tabular}{l|ccccc}
\toprule
 \multicolumn{1}{l|}{\multirow{2}{*} {Model}} & \multicolumn{5}{c}{WebQSP} \\
\cline{2-6}
 \multicolumn{1}{c|}{} &F1&Hits@1& Acc & EM & BM  \\ 
\hline
\hline
\rowcolor{gray!10} \model & \textbf{81.2} & \textbf{84.3} & \textbf{75.2} & \textbf{63.9} & 76.4\\
w/o \textit{SN} & 80.1 & 83.3 & 74.3& 63.6 & \textbf{77.0}\\
w/o \textit{SA} & 79.5 & 82.6 & 73.8 & 63.0 &75.7 \\
w/o \textit{RG} &79.4 & 82.2 &74.3 & 63.4 & 76.7\\
w/o \textit{SA\&RG} & 78.7& 81.0 & 72.5  & 62.1 & 73.6 \\
w/o \textit{SA\&SN}	&79.1	&82.0	&73.8&	63.1&	76.2\\
w/o \textit{ALL} & 76.2 & 79.5 & 70.1 & 59.7 & 72.4  \\
\bottomrule
\end{tabular}
 \caption{Ablation study of sub-modules on WebQSP dataset.}
\label{tab:ablation}
\end{table}

\subsection{Ablation Study}
In this section, we conduct a series of ablation studies to address the following question: 

\textbf{How do the proposed modules improve performance?} Specifically, we conduct experiments against five variants:
1) w/o \textit{SN}: without siamese network, and relevance score is calculated by vector inner product; 2) w/o \textit{SA}: without self-alignment module; 3)  w/o \textit{RG}: without relevance gating module; 4) w/o \textit{SA\&RG}: without both self-alignment and relevance gating modules, where \model obtains prompt embeddings with only MLP. 5) w/o \textit{ALL}: directly appending the retrieval knowledge as context instead of converting retrieval knowledge to prompt embeddings.
As shown in Table \ref{tab:ablation}, we observe that the performance on most metrics decreases when either \textit{SN}, \textit{SA}, or \textit{RG} is removed. This validates the effectiveness of the proposed sub-modules. Furthermore, we find that performance significantly drops when the entire framework is removed, indicating that appending retrieved knowledge directly as context text introduces a large amount of noise, preventing LLMs from focusing on learning the mapping from question to logical form.
Additionally, we notice that the BM is higher in \model w/o \textit{SN} than in \model. This suggests that the ground-truth logical forms are mostly ranked within the top 2 or lower positions during beam search generation, leading to a lower EM.

\begin{table}[t]
\renewcommand{\arraystretch}{1.1} %可以控制行间距
\footnotesize
\centering
\fontsize{9}{11}\selectfont
\setlength{\tabcolsep}{2.8mm}
\begin{tabular}{l|ccccc}
\toprule
 \multicolumn{1}{l|}{\multirow{2}{*} {Model}} & \multicolumn{5}{c}{WebQSP} \\
\cline{2-6}
 \multicolumn{1}{c|}{} &F1&Hits@1& Acc & EM & BM  \\ 
\hline
\hline
\rowcolor{gray!10} \model & \textbf{81.2} & \textbf{84.3} & \textbf{75.2} & \textbf{63.9} & \textbf{76.4} \\
w/o \textit{Relation} & 78.6 & 81.7 &73.1 & 63.0& 74.9\\
w/o \textit{Entity} & 79.2 & 82.2& 73.5& 63.8 & 74.4\\
w/o \textit{Subgraph} &79.7 &82.9 &73.8 & 63.5 & 75.1 \\
\bottomrule
\end{tabular}
 \caption{Quantitative comparison of the impacts of retrieval information on \model's performance.}
\label{tab:ablation retrieval}
\end{table}

\textbf{What impacts do different aspects of retrieval information have on performance?}
To preserve the integrity of \textsc{Amar}, we remove retrieval knowledge by replacing the text embedding with randomly initialized ones. As shown in Table \ref{tab:ablation retrieval}, the results indicate obvious performance drops after removing retrieval knowledge, including `\textit{subgraph}', `\textit{entity}', and `\textit{relation}'. This decline highlights the significance of different aspects of the retrieval knowledge on the overall performance.
Further analysis reveals that removing the `\textit{relation}' component results in the largest drop in performance, suggesting that `\textit{relation}' plays a crucial role in generating logical expressions. Instead, while the `\textit{subgraph}' still contributes to performance, it appears to be less critical for logical forms than `\textit{relation}' or `\textit{entity}'. These findings provide valuable insights for further optimization of the model.

\subsection{Number of Retrieval Analysis}
\begin{figure}[t]
\centering
\begin{minipage}[c]{0.225\textwidth}
    \centering
    \includegraphics[width=\textwidth]{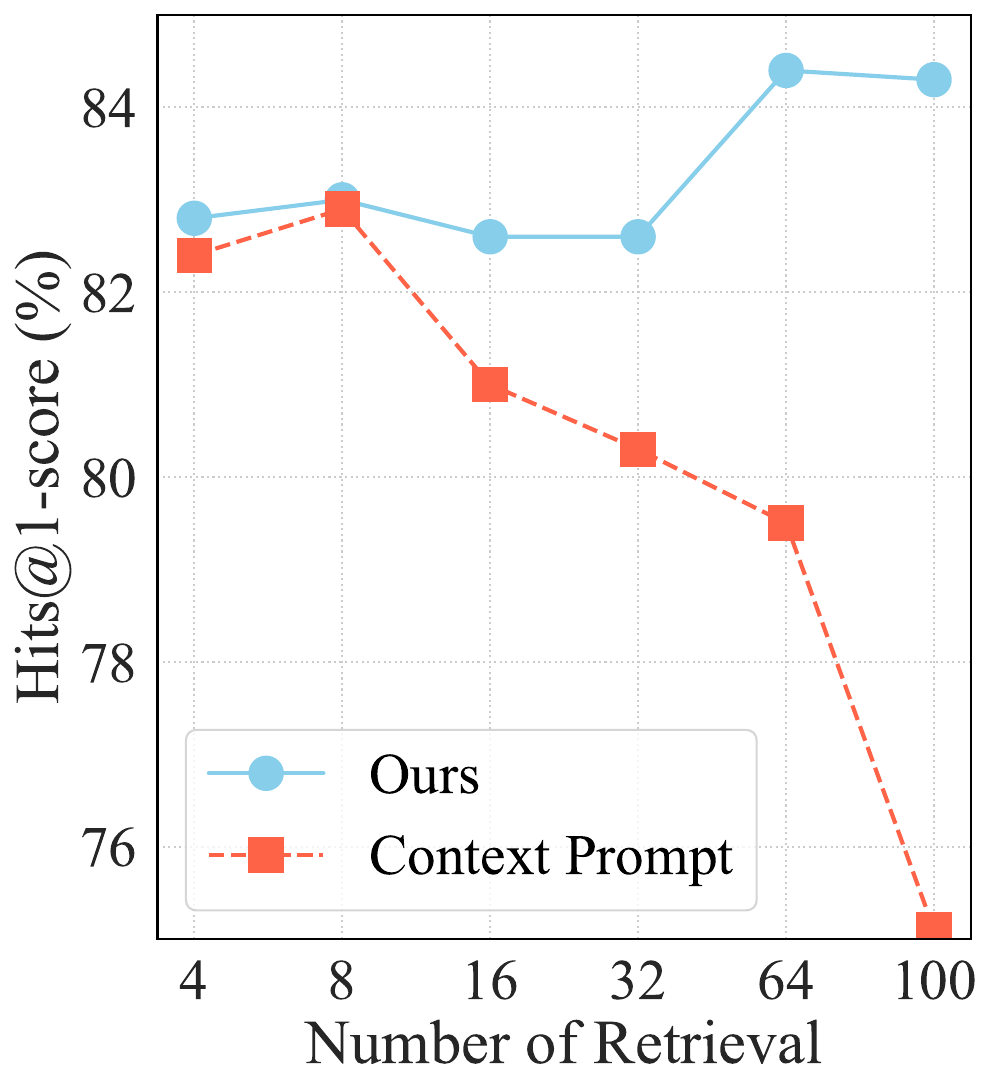}
    \caption{Performance vary with the number of Retrieval.}
    \label{fig:topk}
\end{minipage}
\begin{minipage}[c]{0.225\textwidth}
    \centering
    \includegraphics[width=\textwidth]{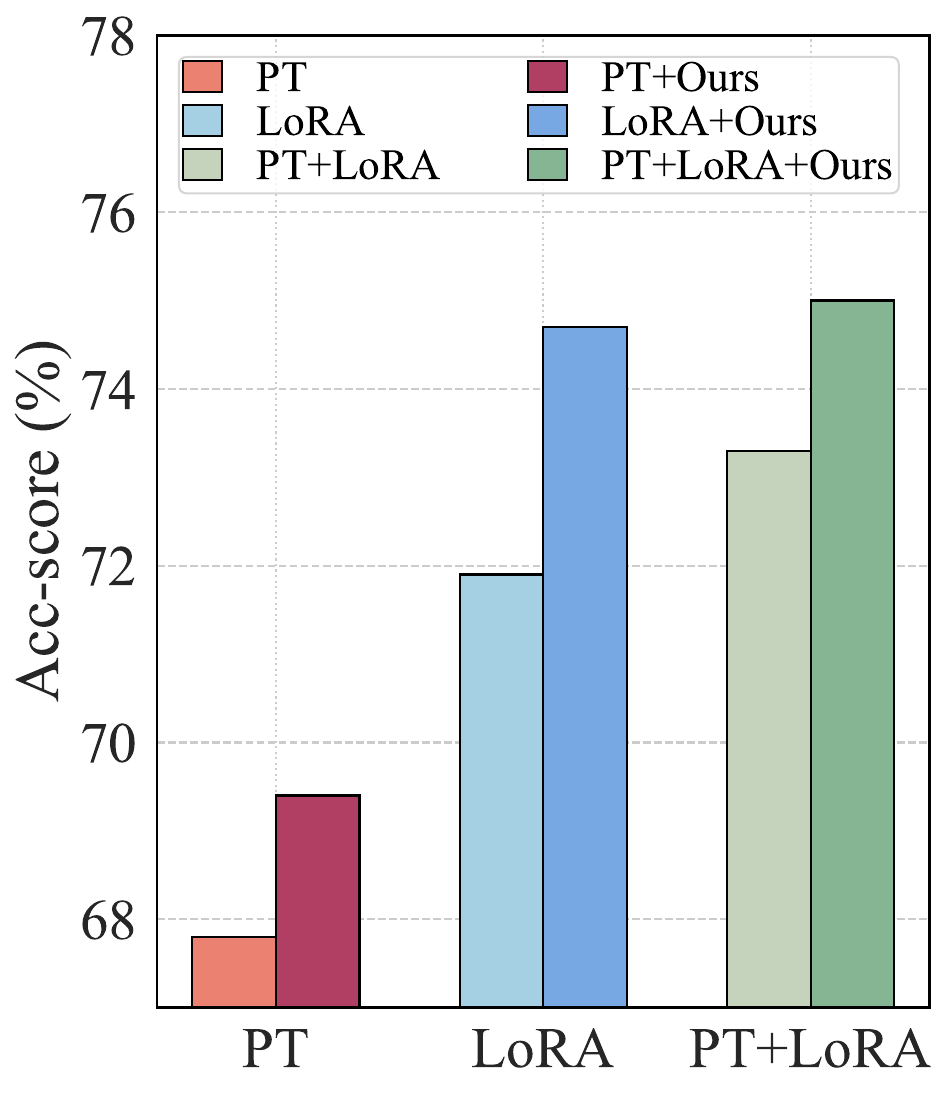}
    \caption{Combine with different fine-tuning methods.}
    \label{fig:Tuning}
\end{minipage} 
\end{figure}

In this section, we explore the impact of the number of retrieval knowledge. We compare the approach of directly inputting retrieved data as a \textit{context prompt}. If the input exceeds the maximum context limit (i.e., 4096 for LLaMA2), we truncate the retrieved information from subgraphs. As shown in figure \ref{fig:topk}, it can be observed that when the amount of retrieved data is relatively small, our method does not significantly differ from  \textit{Context Prompt}, which suggests that useful information recalled is still limited.
However, as the quantity of retrieved data increases (e.g., reaching 64 or 100), our method achieves a substantial performance improvement, while \textit{context prompt} drastically declines. This demonstrates that introducing a long context results in substantial noise, making it difficult for LLMs to learn important data.
In contrast, by treating retrieved information as an individual prompt embedding, we avoid the issue of excessively long inputs and better utilize the rich information.

\begin{table*}[t]
\centering
\footnotesize
\begin{tabular}{p{2.7cm}p{14cm}}
\toprule
Question & \textit{what highschool did harper lee go to?} \\
\midrule 
Entity Retrieval & \textit{\ding{172} Harper Lee {[0.9272]}, \ding{173} Senior secondary education {[0.5441]}, 
 \ding{174} Lee Remick {[0.6482]}, \ding{175} Secondary education {[0.5412]}, \ding{176} Barbara Kingsolver {[0.4320]}, \ding{177} High school movement {[0.6758]}}  \\
 \midrule 
Relation Retrieval &  \textit{\ding{172}  people.person.education {[0.9844]},  \ding{173} education.education.institution {[0.9844]},  \ding{174} education.educational\_institution.school\_type {[0.7281]}, \ding{175} education.school.lowest\_grade\_taught {[0.5469]},  \ding{176} education.school\_mascot.school {[0.5431]}, \ding{177} common.topic.notable\_types {[0.8252]}} \\
\midrule 
 Logical Form by \model & \textit{(AND (JOIN common.topic.notable\_types High school) (JOIN (R education.education.institution) (JOIN (R people.person.education) Harper Lee)))} { \CheckmarkBold} \\
\midrule 
 Logical Form by Context Prompt  & \textit{(AND (JOIN \underline{education.educational.institution.school\_type} \underline{School}) (JOIN (R education.education.institution) ( JOIN (R people.person.education) Harper Lee)))} { \XSolidBrush} \\
\bottomrule 
\end{tabular}
\caption{A case study on WebQSP, where the `\textit{[float]}' represents the scores assigned to each retrieval information, indicating the level of influence it has on the model and the \underline{text with underline} means erroneous generation.
}
\label{tab:case}
\end{table*}

\subsection{Efficiency of Fine-Tuning}

In this section, we analyze the efficiency of our method combined with different fine-tuning methods, including Prompt Tuning (PT)~\cite{lester2021prompttuning} and Low-Rank Adaptation (LoRA). To ensure fairness, we conduct fine-tuning experiments without our module by concatenating the retrieved knowledge text with the input context. As shown in Figure \ref{fig:Tuning}, we observe a significant improvement in performance after incorporating \textsc{Amar}, Regardless of whether PT or LoRA is used, our method consistently outperforms baselines.  Notably, the combination of our method with PT+LoRA fine-tuning yields the best results. This highlights the capability of \model to effectively learn from retrieved information, while the direct concatenation of context introduces considerable noise.
Furthermore, we find that LoRA fine-tuning outperforms PT fine-tuning. This can be attributed to the inherent complexity of logical form generation in the KGQA task.  
LoRA has more tunable parameters and can act on all project layers, thus enabling LLMs to better adapt to tasks.

\subsection{Case Study}

In this section, we present a case study to illustrate how our model adaptively learns the importance of retrieval knowledge. We have not presented a case for subgraph retrieval due to its extensive length. As shown in Table \ref{tab:case}, our method is proven effective in assigning high scores to correct entities and relations for entity retrieval and relation retrieval, while irrelevant or misleading information receives low scores. For instance, the entity `\textit{Harper Lee}' received a score of [0.9272], and relation `\textit{people.person.education}' received a score of [0.9844].
Nevertheless, when retrieved text is directly used as the context prompt, it is susceptible to interference from erroneous information in retrieved data. This can result in generating incorrect relations, such as `\textit{education.educational.institution.school type}'. This example highlights that our method not only enhances the performance of LLM in KGQA but also improves the quality of retrieved information by setting weight.

\subsection{Analysis of LLM Backbones}
In this section, we investigate the question: \textbf{Does the performance improvement of our method solely come from LLMs?} To answer this, we conduct experiments and compare results using different LLMs as backbones: LLaMA2-7B with fine-tuning, LLaMA2-13B with fine-tuning, and GPT-4 (frozen).
% We observe that when using both LLama2-7b and 13b as backbones, \model outperforms baselines using the same backbone, even better than ToG-R~\cite{TOG} using GPT-4.
In Table \ref{tab:llms}, we observe that \model with LLaMA2-7B and 13B as backbones outperform the baselines using the same backbone, such as ChatKBQA~\cite{chatkbqa}, and both even surpass ToG-R~\cite{TOG} using GPT-4 as the backbone on two datasets.
% Moreover, experiments with LLaMA2-7B even reach slightly better results than with LlaMA2-13B on WebQSP because of over-fitting.
These results suggest that the performance gain of \model is not solely attributed to the use of more capable LLMs but rather to the proposed utilization of the commonality between multi-aspect knowledge and the relevance of the question.
We found that LLaMA2-13B perform worse than LLaMA2-7B on WebQSP. We believe the reason lies in dataset characteristics: the scale of WebQSP (1) is much smaller than that of CWQ, and (2) WebQSP has a maximum complexity, only consisting of 2-hop questions. This may cause the LLaMA2-13B to overfit, leading to reduced performance.

\begin{table}[t]
\fontsize{9}{11}\selectfont
\centering
\setlength{\tabcolsep}{3.7mm}{
\begin{tabular}{l|cccc}
\toprule
 \multicolumn{1}{l|}{\multirow{2}{*} {Model}}   & \multicolumn{2}{c}{WebQSP}  & \multicolumn{2}{c}{CWQ}     \\
\cline{2-5}
 \multicolumn{1}{c|}{}   &Acc &Hits@1 &Acc&Hits@1 \\ 
  \hline
  \multicolumn{5}{c}{\textit{GPT-4}} \\
\hline
 ToG-R &- &82.6 &-  & 69.5\\
\hline
\multicolumn{5}{c}{\textit{LLaMA2-7B}} \\
\hline
G-Retriever & - & 70.1& - & - \\
GNN-RAG  &-  & 82.8& -& 62.8\\
ChatKBQA & 73.8 & 83.2 & 73.0 & 82.3 \\
\rowcolor{gray!10}  \model & \textbf{75.2} &\textbf{84.3} & 73.4 & 82.9 \\
 \hline
 \multicolumn{5}{c}{\textit{LLaMA2-13B}} \\
\hline
ChatKBQA & 73.1 & 82.7 & 73.3 & 82.7 \\
\rowcolor{gray!10}  \model & 74.7  &83.3 &\textbf{ 74.5} & \textbf{83.1}\\
\bottomrule
\end{tabular}}
 \caption{Analysis on different LLM backbones. }
\label{tab:llms}
\end{table}

% Answer : Monroe County High School

\section{Related Work} \label{sec:relatedwork}

\noindent
\textbf{Knowledge Graph Question Answering (KGQA).}
KGQA aims to answer questions over KG, and previous methods are usually categorized as EM-based, IR-based, SP-based and LLM-based methods.
EM-based methods encode the entities and relations in the embedding space and reason final answer using these structural embeddings~\cite{shi2021transfernet}.
% For instance, QA-GNN~\cite{yasunaga2021qa} utilizes graph neural networks (GNNs) to reason both KG subgraphs and QA contexts jointly.
IR-based KBQA methods propose to retrieve and re-rank answers from KGs given information conveyed in the question~\cite{chen2019uhop,zhang2022subgraph}.
SP-based methods, focus on transforming question into a structural query, such as SPARQL and S-expression, and reason final answers using these executable queries~\cite{liang2017neural,lan2020query}.
Besides, recent attempts have been made to utilize LLM-based methods for KGQA~\cite{jiang2023unikgqa,chenplan}.
For instance, \citet{RoG} presents a planning-retrieval-reasoning pipeline. ToG \cite{TOG} proposes to interactively explore paths and reasoning on KGs using LLM as an agent.
% G-Retriever is a retrieval-augmented method for general textual graphs and it integrates graph neural networks and LLMs via soft prompting~\cite{G-retriever}.
Despite their significant improvements, a substantial challenge persists when handling multi-aspect retrieved data as input, which may introduce irrelevant knowledge.
However, our method is capable of aligning knowledge and implementing adaptive relevance gating with questions, thus addressing this issue.

% \noindent
% \textbf{Parameter-Efficient Fine-Tuning (PEFT).}
% % 参考 G-Retriever， 简单介绍 lora  和 prompt tuning 等方法。
% Fine-tuning (FT) can provide additional knowledge to LLMs and adjust LLMs' input and output to, for example, adapt to the structured data. Various parameter-efficient fine-tuning (PEFT) methods have been proposed to refine LLMs with minimal training costs~\cite{hu2022lora,zhang2023adalora,li2021prefix,lester2021prompttuning}.
% Low-Rank Adaptation (LoRA) reduces the number of trainable parameters by decomposing a large matrix into two smaller low-rank matrices in the attention layers~\cite{hu2022lora}, and AdaLoRA improves on LoRA by adaptively allocating the parameters among different weight matrices and layers according to their importance scores~\cite{zhang2023adalora}.
% Prompt tuning adds task-specific prompts to the tuning data and these prompt parameters are updated during fine-tuning~\cite{lester2021prompttuning}.
% Prefix tuning is proposed to prefix task-specific vectors to the tuning data and only updates these prefix parameters while keeping the LLM's parameters frozen~\cite{li2021prefix}.

% However, PEFT methods, such as LoRA and Prompt tuning, may face difficulties in effectively utilizing the extensive information derived from long texts, which often yield subpar model performance \cite{chen2023longlora}. In contrast, our model designed an adaptive framework for the retrieval information, enabling efficient learning the mapping between knowledge to generate the desired output.

\noindent
\textbf{Large Language Model Reasoning.}
Considering the impressive abilities of LLMs, some previous works focus on facilitating LLMs' reasoning via prompting~\cite{NSM,CBR-KBQA,structgpt,wang2023knowledge,li2024visualization,10.1007/978-981-97-7232-2_18,chen2024tackling}.
% For example, as a neuro-symbolic approach, CBR-KBQA reuses existing cases to improve reasoning on unseen cases~\cite{CBR-KBQA}.
To overcome the unfaithful reasoning of LLMs, ~\citet{structgpt} proposes StructGPT, an iterative reading-then-reasoning framework, to improve reasoning of LLMs when handling structured data.
 % FAME leverages Monte-Carlo planning to generate faithful reasoning steps~\cite{hong2023faithful} and
KD-CoT is designed to formulate chain-of-thought into a multi-round QA format and LLMs can retrieve external knowledge during interaction~\cite{wang2023knowledge}.
% ~\citet{du2024improving} proposes to utilize multiple LLMs and obtain the final answer through discussion and debation among these LLMs, which improves the factual validity of the generated contents.
However, LLMs reasoning is challenging when leveraging information present in lengthy texts \cite{wang2024beyond}. In contrast, our model facilitates efficient selection of information by utilizing prompt embeddings, thereby mitigating the issue of excessive context.

\section{Conclusion}
In this work, we propose a novel approach to enhance LLMs reasoning and factual output by retrieving multi-aspect knowledge from KGs. By employing self-alignment and relevance gating modules, \model adaptively enhances and selects relevant information. 
It has proven more effective than simply appending retrieved text to input context, as it minimizes noise interference. Through extensive experiments, \model outperforms 22 baseline models and achieves a new state-of-the-art performance.

\section{Acknowledgment}
This work was supported in part by the grants from National Science and Technology Major Project (No. 2023ZD0121104), and National Natural Science Foundation of China (No.62222213, 62072423).
This research was partially supported by Research Impact Fund (No.R1015-23), APRC - CityU New Research Initiatives (No.9610565, Start-up Grant for New Faculty of CityU), CityU - HKIDS Early Career Research Grant (No.9360163), Hong Kong ITC Innovation and Technology Fund Midstream Research Programme for Universities Project (No.ITS/034/22MS), Hong Kong Environmental and Conservation Fund (No. 88/2022), and SIRG - CityU Strategic Interdisciplinary Research Grant (No.7020046), Tencent (CCF-Tencent Open Fund, Tencent Rhino-Bird Focused Research Program).
\bibliography{aaai25}

\newpage
\section{Appendix}

\subsection{Comparison in Oracle Entities setting}

In this section, we present the results of our experiments conducted under the Oracle Entity Linking Annotations setting. This setting assumes that the model has prior knowledge of the correct entity mentioned in the question. The outcomes of this experiment can be seen in Table \ref{tab:orcale setting result}.
The results of our experiment indicate that our method achieves state-of-the-art (SOTA) improvements. Interestingly, our approach performs exceptionally well even when the Oracle setting is not utilized. In fact, our method outperforms several baseline models that rely on the Oracle setting.
For example, when evaluating the WebQSP F1 metric, RoG obtains a score of 70.8. However, our method surpasses this score significantly, achieving a score of 81.2. This remarkable achievement demonstrates the effectiveness of our approach.
\begin{table*}[t]
% \setlength\tabcolsep{3pt}  %可以控制列间距
% \renewcommand{\arraystretch}{1.1} %可以控制行间距
% \footnotesize
\centering
% \fontsize{9}{11}\selectfont
\setlength{\tabcolsep}{2.6mm}{
\begin{tabular}{l|cccccc}
\toprule
  \multicolumn{1}{l|}{\multirow{2}{*} {Model}}   & \multicolumn{3}{c}{WebQSP}   & \multicolumn{3}{c}{CWQ} \\
\cline{2-7}
 \multicolumn{1}{c|}{}   &F1&Hits@1& Acc &F1&Hits@1& Acc\\ 
\hline
\hline
TIARA* \cite{shu-etal-2022-tiara} & 78.9& 75.2& - &- &- &- \\
EmbedKGQA* & - & 66.6& - & - & 45.9& -  \\
ProgramTransfer* \cite{ProgramTransfer}& 76.5& 74.6& - &58.7& 58.1 &- \\
ChatKBQA* \cite{chatkbqa}& 83.5 &86.4 &77.8 &81.3 &86.0 &76.8\\
RoG* \cite{RoG}& 70.8 &85.7& -  &56.2& 62.6& - \\
FiDeLiS-GPT4* \cite{sui2024fidelis}&  78.32&  84.39&  -&  64.32&  71.47&  - \\
\rowcolor{gray!10}  \model (Ours) & 81.2\scriptsize{$\pm$0.15} & 84.3\scriptsize{$\pm$0.16}  & 75.2\scriptsize{$\pm$0.10} & 78.5\scriptsize{$\pm$0.11} & 83.1\scriptsize{$\pm$0.09} & 74.5\scriptsize{$\pm$0.07} \\
\rowcolor{gray!10}  \model* (Ours) & \textbf{84.1}\scriptsize{$\pm$0.13} & \textbf{87.0}\scriptsize{$\pm$0.11} & \textbf{78.4}\scriptsize{$\pm$0.09}  & \textbf{82.0}\scriptsize{$\pm$0.10}& \textbf{86.4}\scriptsize{$\pm$0.11} &  \textbf{78.2}\scriptsize{$\pm$0.09} \\
\bottomrule
\end{tabular}}
\caption{Performance comparison of KGQA methods in oracle entity linking annotations setting. \textbf{*} means methods with oracle entity.}
\label{tab:orcale setting result}
\end{table*}

\subsection{Logical Form}
In this study, we have chosen the S-expression $\mathcal{F}$ as our logical expression, which has been previously used in \cite{chatkbqa,decaf}. The S-expression, as exemplified by `\textit{(AND (JOIN base.biblioness.bibs\_location.loc\_type "Country") (JOIN (R location.location.contains) Oceania))}', utilizes functions such as \textit{JOIN} and \textit{AND} to operate on set-based semantics. This approach strikes a balance between readability and compactness, making it suitable for Knowledge Graph Question Answering (KGQA) \cite{gu2021beyond}.
The \textit{JOIN} operation is used to query a triple (h, r, t) on either h or t. For instance, (?, r, t) is denoted as \textit{(JOIN r t)}, and (h, r, ?) is denoted as \textit{(JOIN (R r) h)}.
E1 or E2 denote a sublayer logical form. Various operators include:
\begin{itemize}
    \item \textit{`AND' (AND E1 E2)}: represents the intersection of E1 and E2.
    \item \textit{`COUNT' (COUNT E1)}: denotes the count of E1.
    \item \textit{`ARGMAX' (ARGMAX E1 r)}: represents the maximum literal obtained after projecting E1 onto the r relation.
    \item \textit{`ARGMIN' (ARGMIN E1 r)}: denotes the minimum literal obtained after projecting E1 onto the r relation.
    \item \textit{`GT' (GT E1 l)}: represents the portion of E1 that is greater than l.
    \item \textit{`GE' (GE E1 l)}: denotes the part of E1 that is greater than or equal to l.
    \item \textit{`LT' (LT E1 l)}: represents the part of E1 that is less than l.
    \item \textit{`LE' (LE E1 l)}: denotes the part of E1 that is less than or equal to l.
\end{itemize}

\subsection{Retrieval of Multi-Aspect Knowledge}
\textbf{Entity Retrieval.} 
One effective approach for retrieving candidate entities $k_e$ is to conduct entity linking with question $q$. Following \citet{GMT-KBQA}, we employ the ELQ \cite{ELQ} for question entity linking. ELQ is an efficient and precise entity linking system designed specifically for questions. The system aim to identify the boundaries of entity mentions within a question and link them to their corresponding Wikipedia entities. The system uses a bi-encoder based on BERT \cite{devlin2018bert} to achieve this. The entity encoder calculates entity embeddings for all entities in Wikipedia \cite{wikidata} using their short descriptions. Simultaneously, the question encoder generates token-level embeddings for the input question. These embeddings are then used to detect mention boundaries and disambiguate each entity mention. This is done by calculating an inner product between the mention embeddings (which are an average of the mention tokens) and the entity embeddings.
Then FACC1 \cite{facc1} (a comprehensive Freebase annotation of corpora) is employed to identify entities that were not linked by ELQ, to enhance the range of candidate entities. The entity recall score on WebQSP and CWQ datasets are shown in Table \ref{tab:Recall entity}.

\textbf{Relation Retrieval.}
As we have stated in main body, in large-scale KG (e.g. Freebase), relations are typically organized hierarchically. Therefore, directly using question-based dense retrieval for similarity may not be effective. To address this, we propose masking entity mentions detected during the candidate entity retrieval stage with a [BLANK] token for each question $q$. For example, `\textit{(AND (JOIN base.biblioness.bibs\_location.loc\_type Country) (JOIN (R location.location.contains) Oceania))}' is replaced by `\textit{(AND (JOIN base.biblioness.bibs\_location.loc\_type [BLANK]) (JOIN (R location.location.contains) [BLANK]))}'. Following \citet{GMT-KBQA,CBR-KBQA}, we train two separate BERT \cite{devlin2018bert} models that encode questions $q$ and relations $r$ into a shared dense space. And we calculate the similarity score by dot-product:
 \begin{equation}
\begin{aligned}
    &\boldsymbol{v}_{q} = BERT(q), \\
   &\boldsymbol{v}_{r} = BERT(r), \\
    &\boldsymbol{s}(q,r) = \boldsymbol{v}_{q} \cdot \boldsymbol{v}_{r}. \\
\end{aligned}
\end{equation}
To construct a training batch, we randomly select negative relations that are not part of the logical form of a given question. The objective of optimization is to maximize the score of the relevant relation compared to the randomly sampled relations. 
 To retrieve the nearest relations, we employ FAISS \cite{douze2024faiss}, a highly efficient vector database, which allows us to speed up the search process and obtain the most relevant results. 
Next, we proceed to train a ranker that assigns scores to question and relation pairs. To achieve this, we utilize a cross-encoder, which is a single BERT model. The input to cross-encoder is a combination of question and candidate relation. By employing a linear layer, we project the representation of combined input ([$q$; $r$]) to a binary probability distribution. This allows us to calculate score between $q$ and $r$. During training, we employ cross-entropy loss to optimize this process. Finally, we retain the top-k candidate relations based on their rankings.
The relation recall score on WebQSP and CWQ datasets are shown in Table \ref{tab:Recall relation}.

\textbf{Subgraph Retrieval.}
To better utilize structural and semantic information contained within KG, we linearize triplets by combining the head entity, relation, and tail entity with spaces for retrieval. Drawing inspiration from \cite{decaf}, we propose grouping linearized sentences with the same head entity into a document. To save computing resources, we only focus on 1-hop subgraphs to capture structural information. For example, the triple \textit{(Oceania, location.location.contains, Australia)} can be linearized to `\textit{Oceania location location contains Australia}'.  All linearized texts from triplet connected to the same entity will be concatenated together. Each document is truncated with a maximum of 100 words.
Furthermore, concerning the potential information loss when converting long documents into vectors, we employ sparse retrieval approaches that rely on keyword dependencies. Specifically, we employ techniques like BM25 \cite{robertson2009probabilistic}, The BM25 algorithm is widely employed for scoring search relevance. In essence, it involves analyzing the query to generate morphemes $q_i$ through morphological analysis. For each search result $D$, the algorithm computes the relevance score between each morpheme $q_i$ and $D$. These relevance scores are then weighted and combined to determine the overall relevance score between the query and $D$.
 We follow BM25 to calculates TF-IDF scores based on sparse word matches between input questions and KB-linearized passages, and obtain the fine rerieved subgraphs.
The subgraph recall score on WebQSP and CWQ datasets are shown in Table \ref{tab:Recall subgraph}.

\begin{table}[t]
\centering
\begin{tabular}{lcc}
\toprule
Retrieved Entities & WebQSP  & CWQ  \\
\midrule 
Top 1 &67.7 &48.4 \\
Top 2 &76.1 &72.0 \\
Top 3 &77.8 &76.9 \\
Top 4 &78.8 &78.2 \\
Top 5 &79.4 &78.9 \\
Top 6 &80.0 &79.3 \\
Top 7 &80.1 &79.6 \\
Top 8 &80.3 &79.9 \\
Top 9 &80.3 &80.0 \\
Top 10 &80.6 &80.2 \\
\bottomrule 
\end{tabular}
\caption{Recall score (\%) on WebQSP and CWQ datasets. This metric measures the recall of groundtruth entities in the retrieved entities information.
}
\label{tab:Recall entity}
\end{table}

\begin{table}[t]
\centering
\begin{tabular}{lcc}
\toprule
Retrieved Relations & WebQSP  & CWQ  \\
\midrule 
Top 1& 41.4& 29.1 \\
Top 2& 63.1& 54.1 \\
Top 3& 75.7& 71.0\\
Top 4& 82.3& 80.2\\
Top 5& 86.0& 85.4\\
Top 6& 88.1& 87.9\\
Top 7& 89.3& 89.5\\
Top 8& 90.3& 90.6\\
Top 9& 91.2& 91.4\\
Top 10& 92.0& 92.0\\
\bottomrule 
\end{tabular}
\caption{Recall score (\%) on WebQSP and CWQ datasets. This metric measures the recall of groundtruth relations in the retrieved relations information.
}
\label{tab:Recall relation}
\end{table}

\begin{table}[t]
\centering
\begin{tabular}{lcc}
\toprule
Retrieved Subgraphs & WebQSP  & CWQ  \\
\midrule 
Top 5 &30.7 &27.9  \\
Top 10 &39.8 &34.8  \\
Top 20 &48.5 &41.0  \\
Top 100 &68.4 &57.4  \\
\bottomrule 
\end{tabular}
\caption{Recall score (\%) on WebQSP and CWQ datasets. This metric measures the recall answers in the retrieved subgraph information.
}
\label{tab:Recall subgraph}
\end{table}

\begin{table*}[htbp]
\centering
% \footnotesize
\begin{tabular}{lrrrrrccc}
\toprule
\multicolumn{1}{l}{\multirow{2}{*} {Dataset}}  & \multicolumn{1}{l}{\multirow{2}{*} {\#Question }} & \multicolumn{1}{l}{\multirow{2}{*} {\#Skeleton}} & \multicolumn{1}{l}{\multirow{2}{*} {\#Train}} & \multicolumn{1}{l}{\multirow{2}{*} {\#Valid}} & \multicolumn{1}{l}{\multirow{2}{*} {\#Test }} & \multicolumn{3}{c}{\#Average Token Length}  \\
 \multicolumn{6}{c}{} & $k_e$ & $k_r$ & $k_s$ \\
\midrule 
WebQSP  & 4,737 &34 &3,098 &- &1,639 & 11.4 & 4.2 & 145.9 \\
CWQ &34,689 &174 &27,639 &3,519 &3,531& 11.0 & 4.1 &147.4 \\
\bottomrule 
\end{tabular}
\caption{Dataset statistics.}
\label{tab:all Statistics}
\end{table*}

\subsection{Query Execution}
After generating S-expressions using LLMs, we need to refine them further. Let's take the S-expression \textit{(AND (JOIN base.biblioness.bibs\_location.loc\_type Country) (JOIN (R location.location.contains) Oceania))} as an example. In this case, our refined targets consist of entities such as \textit{Oceania} and \textit{Country}, as well as relations like \textit{base.biblioness.bibs\_location.loc\_type} and \textit{location.location.contains}. It's important to note that the logical structure of the S-expression remains unchanged.

To refine the S-expressions, we follow the approach proposed by \citet{chatkbqa}. Firstly, we extract the entire set of entities from the KG. Then, we employ unsupervised technique SimCSE \cite{gao2021simcse} to calculate similarity scores between the extracted entities. By applying a certain threshold, we obtain a subset of entities that have a similarity score above the threshold.
Regarding relations, we extract all connected relation neighbors from the subset of entities. Also, we utilize SimCSE to calculate similarity scores, enabling us to identify the candidate relation with the highest similarity score. All settings of executing query follow \citet{chatkbqa}.

After refining the S-expression, we directly convert it into a SPARQL expression. For example, the SPARQL expression for the above example is as follows:
\lstset{language=C}
\begin{lstlisting}
"PREFIX ns: http://rdf.freebase.com/ns/
SELECT DISTINCT ?x
WHERE 
{
FILTER (?x != ns:m.05nrg)
FILTER (!isLiteral(?x) OR lang(?x) = '' OR langMatches(lang(?x), 'en'))
ns:m.05nrg ns:location.location.contains ?x .
?x ns:base.biblioness.bibs_location.loc_type ?sk0 .
FILTER (str(?sk0) = "Country")
}"
\end{lstlisting}
We then use SPARQL expression to query KG, obtaining final results. This process of refining and querying allows us to ensure accuracy and relevance of information retrieved from KG, thereby enhancing the effectiveness of our models.

\subsection{Experiment Settings}
\subsubsection{Datasets}
In our work, we utilized two well-known and commonly used datasets: the WebQuestions Semantic Parses Dataset (WebQSP) \cite{webqsp} and ComplexWebQuestions (CWQ) \cite{CWQ}. The statistics for these datasets are presented in table \ref{tab:all Statistics}. $k_e$, $k_r$ and $k_s$ denote the average token length of entities, relations and subgraphs in the retrieval knowledge.  Both of these datasets are based on Freebase \cite{bollacker2008freebase} as the KG database.

The WebQSP dataset contains full semantic parses in SPARQL queries for 4,737 questions, and partial annotations for the remaining 1,073 questions for which a valid parse could not be formulated or where the question itself is bad or needs a descriptive answer.

The CWQ dataset is a dataset designed for answering complex questions that require reasoning over multiple web snippets. It contains a large set of complex questions in natural language.

\subsubsection{Baselines}

In this study, we compare the performance of \model against 22 baselines, including 4 embedding-based (EM-based) KGQA methods, 5 information retrieval-based (IR-based) KGQA methods, 6 semantic parsing-based (SP-based) KGQA methods, and 7 LLM-based KGQA methods.
In the comparative experiments in the Oracle entities setting, we compare \model against another 6 baselines.
The details of each baseline are given as follows.

\textbf{EM-based KGQA methods}.
KV-Mem~\cite{miller2016key} firstly stores the facts in a key-value structured memory, utilizes different encodings on the reading operation, and reasons on these facts to obtain the answer.
NSM$_{+h}$~\cite{NSM} proposes a teacher-student framework for multi-hop KGQA, in which the student network aims to find the answers using a neural state machine (NSM) and the teacher network tries to learn intermediate supervision signals for enhance the student work.
TransferNet~\cite{shi2021transfernet} begins with the topic entity (within the question) and obtains the answer by transferring entity scores along relation scores of multiple steps.
KGT5~\cite{KGT5} unifies knowledge graph completion and KGQA as the seq-to-seq tasks and performs question answering after fine-tuning using QA pairs.

\textbf{IR-based KGQA methods}.
GraftNet~\cite{sun-etal-2018-open} adopts a graph convolutional network (GCN) to operate on both the KG facts and text sentences and extract answers from a question-specific subgraph.
PullNet~\cite{sun-etal-2019-pullnet} involves training a graph convolutional network (GCN) to improve the retrieval process and multi-hop question answering.
SR+NSM~\cite{zhang2022subgraph} proposes a trainable subgraph retriever to improve the retrieval module, and SR+NSM+E2E~\cite{zhang2022subgraph} trains both the retrieval and reasoning modules in an end-to-end manner based on SR+NSM.
UniKGQA~\cite{unikgqa} proposes to unify the retrieval and reasoning in both model architecture and parameter learning for multi-hop KGQA.
EmbedKGQA~\cite{EmbedKGQA} leverages KG embeddings to perform missing link prediction, thus reducing KG sparsity and improving multi-hop question-answering.

\textbf{SP-based KGQA methods}.
CBR-KBQA~\cite{CBR-KBQA} proposes a neuro-symbolic case-based reasoning (CBR) approach and reuses existing cases to improve reasoning on unseen cases.
GMT-KBQA~\cite{GMT-KBQA} improves logical form generation with multi-task learning and better utilization of auxiliary information.
UnifiedSKG~\cite{xie-etal-2022-unifiedskg} unifies a total of 21 structured knowledge grounding tasks into a text-to-text format and improves the performance via multi-task prefix tuning on T5.
RnG-KBQA~\cite{rng-kbqa} enumerates all the candidate logical forms from KG, ranks these candidates using the contrastive ranker, and then obtains the target logical form with the tailored generator.
DecAF~\cite{decaf} proposes to jointly generate both logical forms and direct answers and then combine them to obtain the final answer.
FC-KBQA~\cite{fc-kbqa} introduces a fine-to-coarse composition framework to improve both the generalization and execution abilities of the generated logical forms.
TIARA~\cite{shu-etal-2022-tiara} applies multi-grained retrieval to help language models concentrate on relevant KG contexts, including entities, schemas, and logical forms.
ProgramTransfer~\cite{ProgramTransfer} proposes a two-stage parsing framework and an ontology-guided pruning strategy to program transfer, leveraging program annotations as supervision signals to assist program induction.

\textbf{LLM-based KGQA methods}.
KD-CoT~\cite{wang2023knowledge} introduces formulating chain-of-thought (CoT) into a multi-round QA format and during the interaction, LLMs can retrieve external knowledge for faithful reasoning.
Pangu~\cite{pangu} proposes to leverage LLMs' discriminative abilities rather than generative abilities for performance improvement on complex KGQA.
StructGPT~\cite{structgpt} is an iterative reading-then-reasoning framework to improve LLMs' reasoning when handling structured data, such as KG.
ChatKBQA~\cite{chatkbqa} follows the generate-then-retrieve KGQA framework, which firstly probes LLMs to generate the logical forms and refines the entities and relations within the logical forms.
ToG-R~\cite{TOG} proposes to explore paths and reasoning interactively on KGs using LLM as an agent.
RoG~\cite{RoG} presents a planning-retrieval-reasoning pipeline to generate relation paths, retrieve reasoning paths, and conduct faithful reasoning on these paths.
G-Retriever~\cite{G-retriever} introduces a retrieval-augmented method for general textual graphs and it integrates GNNs, LLMs, and RAG to improve the question-answering abilities via soft prompting of LLMs.
GNN-RAG~\cite{mavromatis2024gnn} focuses on combining the language understanding abilities of LLMs with the reasoning abilities of GNNs in an RAG style.
FiDeLiS~\cite{sui2024fidelis} presents a retrieval-exploration interactive method, which addresses intermediate reasoning steps on KGs and utilizes deductive reasoning capabilities of LLMs to guide the reasoning process in a step-wise and generalizable manner.

\subsubsection{Implementation Details}
Our implementation is based on NVIDIA A6000 GPUs, requiring approximately 48GB of VRAM for training and testing. However, executing querying on KG only requires 3GB of VRAM. Below, we provide the selected hyperparameters for the WebQSP and CWQ datasets. We only searched for the number of retrieved data.

\textbf{For the WebQSP dataset:}
LoRA target weight: gate\_proj down\_proj up\_proj
Learning rate: 5e-5,
LoRA rank: 8,
LoRA alpha: 32,
LoRA dropout: 0.1,
Training epochs: 80,
Training batch size: 4,
Number of retrieved data: \{4, 8, 16, 32, 64, 100\},
Soft prompt length: 7,
Beam search number: 8,
Max new tokens: 256.

\textbf{For the CWQ dataset:}
LoRA target weight: gate\_proj down\_proj up\_proj,
Learning rate: 5e-5,
LoRA rank: 8,
LoRA alpha: 32,
LoRA dropout: 0.1,
Training epochs: 10,
Training batch size: 4,
Number of retrieved data: \{4, 8, 16, 32, 64, 100\},
Soft prompt length: 16,
Beam search number: 15,
Max new tokens: 256.

For more code details, please refer to the source code in the supplementary materials.

\subsubsection{Experiments on training time and space consumption}
These results shown in Table \ref{tab:time}, obtained using an A6000 GPU with LLaMA2-7b on webQSP (batch\_size=4), demonstrate that (1) time and space consumption are within acceptable range; (2) our method significantly reduces both time and space costs compared to directly using text as the context.

\begin{table}[t]
\setlength\tabcolsep{2pt}  %可以控制列间距
\centering
\begin{tabular}{lcc}
\toprule
 & Training Time & GPU Memory  \\
\midrule 
Context Prompt&	7.24 min/epoch	&31 GB \\
Ours	&4.87 min/epoch	&18 GB\\
\bottomrule 
\end{tabular}
\caption{Training time and space consumption.
}
\label{tab:time}
\end{table}

\end{document}